\documentclass[preprint,review,12pt]{elsarticle}

\usepackage{amssymb}
\usepackage{amsthm}
\usepackage{subeqnarray}
\usepackage{cases}
\usepackage{multirow}
\usepackage{subfigure}
\usepackage{graphicx}
\usepackage{epsfig,times}
\usepackage{algorithm,algorithmic}

\newtheorem{theorem}{Theorem}[section]
\newtheorem{proposition}[theorem]{Proposition}
\renewcommand{\d}{\mbox{d}}

\journal{IVC}

\begin{document}

\begin{frontmatter}

%% Title, authors and addresses

%% use the tnoteref command within \title for footnotes;
%% use the tnotetext command for the associated footnote;
%% use the fnref command within \author or \address for footnotes;
%% use the fntext command for the associated footnote;
%% use the corref command within \author for corresponding author footnotes;
%% use the cortext command for the associated footnote;
%% use the ead command for the email address,
%% and the form \ead[url] for the home page:
%%
%% \title{Title\tnoteref{label1}}
%% \tnotetext[label1]{}
%% \author{Name\corref{cor1}\fnref{label2}}
%% \ead{email address}
%% \ead[url]{home page}
%% \fntext[label2]{}
%% \cortext[cor1]{}
%% \address{Address\fnref{label3}}
%% \fntext[label3]{}

\title{Toward Designing Intelligent PDEs for Computer Vision: An Optimal Control Approach}

%% use optional labels to link authors explicitly to addresses:
%% \author[label1,label2]{<author name>}
%% \address[label1]{<address>}
%% \address[label2]{<address>}

\author[a]{Risheng Liu}
\author[b]{Zhouchen Lin}
\author[c]{Wei Zhang}
\author[a]{Kewei Tang}
\author[a]{Zhixun Su}
\address[a]{School of Mathematical Sciences, Dalian University of Technology,
Dalian, China.}
\address[b]{Microsoft Research Asia, Beijing, China, e-mail: zhoulin@microsoft.com.}
\address[c]{Department of Information Engineering, The Chinese University of Hong Kong, China.}

\begin{abstract}
%% Text of abstract
Many computer vision and image processing problems can be posed as
solving partial differential equations (PDEs). However, designing
PDE system usually requires high mathematical skills and good
insight into the problems. In this paper, we consider designing PDEs
for various problems arising in computer vision and image processing
in a lazy manner: \emph{learning PDEs from real data via data-based
optimal control}. We first propose a general intelligent PDE system
which holds the basic translational and rotational invariance rule
for most vision problems. By introducing a PDE-constrained optimal
control framework, it is possible to use the training data resulting
from multiple ways (ground truth, results from other methods, and
manual results from humans) to learn PDEs for different computer
vision tasks. The proposed optimal control based training framework
aims at learning a PDE-based regressor to approximate the unknown
(and usually nonlinear) mapping of different vision tasks. The
experimental results show that the learnt PDEs can solve different
vision problems reasonably well. In particular, we can obtain PDEs
not only for problems that traditional PDEs work well but also for
problems that PDE-based methods have never been tried before, due to
the difficulty in describing those problems in a mathematical way.
\end{abstract}

\begin{keyword}
%% keywords here, in the form: keyword \sep keyword
Optimal Control \sep PDEs \sep Computer Vision \sep Image Processing.
%% MSC codes here, in the form: \MSC code \sep code
%% or \MSC[2008] code \sep code (2000 is the default)

\end{keyword}

\end{frontmatter}

%%
%% Start line numbering here if you want
%%
% \linenumbers

%% main text
\section{Introduction}
\begin{figure}
\centering
\includegraphics[width=0.5\textwidth,
keepaspectratio]{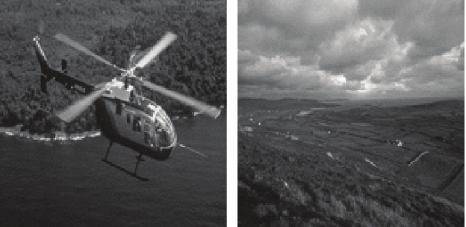}
\caption{Can we design a PDE system which can detect the object of
interest (e.g., the helicopter in the left image) and does not respond
if the object is absent (e.g., the right image)?}\label{fig:intro}
\end{figure}

The wide applications of partial differential
equations (PDEs) in computer vision and image processing can be
attributable to two main factors \cite{Tony:2005:PDEIP}. First, PDEs
in classical mathematical physics are powerful tools to describe,
model, and simulate many dynamics such as heat flow, diffusion, and
wave propagation. Second, many variational problems or their
regularized counterparts can often be effectively solved from their
Euler-Lagrange equations. Therefore, in general there are two types
of methods for designing PDEs for vision tasks. For the first kind
of methods, PDEs are written down directly (e.g., anisotropic
diffusion \cite{Pietro:1990:PM}, shock filter
\cite{Osher:1990:ShockFilter}, based on some understandings on the
properties of mathematical operators and the physical natures of the
problems, and curve-evolution-based equations
\cite{Sapiro:2001:GPDE}). The second kind of methods basically
define an energy functional and then derive the evolution equations
by computing the Euler-Lagrange equation of the energy functional
(e.g., total-variation-based variational methods
\cite{Rudin:1992:ROF}\cite{Chan:2005:TVL1}\cite{David:2003:TVEdge}).
In either way, people have to heavily rely on their intuition on the
vision tasks. Therefore, traditional PDE-based methods require good
mathematical skills when choosing appropriate PDE forms and
predicting the final effect of composing related operators such that
the obtained PDEs roughly meet the goals. If people do not have
enough intuition on a vision task, they may have difficulty in
acquiring effective PDEs. For example, although there has been much
work on PDE-based image segmentation
\cite{Bresson:2007:TVSeg}\cite{Chambolle:2005:TVLevelSet}\cite{Li:2005:LevelSet}\cite{Gao:11:levelset},
the basic philosophy is always to follow the strong edges in the
image and also require the edge contour to be smooth. Can we have a
PDE system for objective detection (Fig.~\ref{fig:intro}) that
locates the object region if the object is present and does not
respond if the object is absent? We believe that this is a big
challenge to human intuition and is much more difficult than
traditional segmentation tasks if a PDE-based method is required,
because it is hard to describe an object class, which may have
significant variation in shape, texture and pose. Without using
additional information to judge the content, the existing PDEs for
segmentation, e.g., \cite{Li:2005:LevelSet}, always output an
``object region'' for any non-constant image. In short, current PDE
design methods greatly limit the applications of PDEs to a wider and
more complex scope. This motivates us to explore whether we can
acquire PDEs that are less artificial yet more powerful. In this
paper, we give an affirmative answer to this question. We
demonstrate that learning particular coefficients of a general
intelligent PDE system from a given training data set might be a
possible way of designing PDEs for computer vision in a lazy manner.
Furthermore, borrowing this learning strategy from machine learning
can generalize PDEs techniques for more complex vision problems.

%\begin{figure}
%\begin{center}
%\begin{tabular}{c@{\extracolsep{0.2em}}c}
%\includegraphics[width=0.15\textwidth,
%keepaspectratio]{image/eps/intro/plane.eps}
%&\includegraphics[width=0.15\textwidth,
%keepaspectratio]{image/eps/intro/scencry.eps}\\
%\end{tabular}
%\end{center}
%\caption{Can we design a PDE system which can detect the object of
%interest (e.g., the helicopter in the left image) and does not respond
%if the object is absent (e.g., the right image)?}\label{fig:intro}
%\end{figure}

Inspired by the electromagnetic field theory and Maxwell's equations
\cite{Cheng:1989:EM}, we assume that the visual processing has two
coupled evolutions in different scale spaces: one is in the
\emph{image scale space}, which controls the evolution of the
output, and the other is in the \emph{indicator scale space} that
helps collect the global information to guide the evolution in the
image scale space. In this way, our general intelligent PDE system
consists of two coupled evolutionary PDEs. Both PDEs are coupled
equations between the image and indicator, up to their second order
partial derivatives. Another key idea of our general intelligent PDE
system is to assume that the PDEs that are sought could be written
as combinations of ``atoms'' which satisfy the general properties of
vision tasks. As a preliminary investigation, we utilize all the
translational and rotational invariants as such``atoms'' and propose
the general intelligent PDE system as a linear combination of all
these invariants \cite{Olver:93:applications}. Then the problem
boils down to determining the combination coefficients among such
``atoms''.

The theory of optimal control \cite{Kirk:1970:Opt} has been well
developed for over fifty years. With the enormous advances in
computing power, optimal control is now widely used in
multi-disciplinary applications such as biological systems,
communication networks and socio-economic systems etc
\cite{Ababnah:11:control}. Optimal design and parameter estimation
of systems governed by PDEs give rise to a class of problems known
as PDE-constrained optimal control \cite{Lions:1971:OptPDE}. In this
paper, a PDE-constrained optimal control technique as the training
tool is introduced for our PDE system. We further propose a general
framework for learning PDEs to accomplish a specific vision task via
PDE-constrained optimal control, where the objective functional is
to minimize the difference between the expected outputs and the
actual outputs of the PDEs, given the input images. Such
input-output image pairs are provided in multiple ways (e.g., ground
truth, results from other methods or manually generated results by
humans) for different tasks. Therefore, we can train the general
intelligent PDE system to solve various vision problems which
traditional PDEs may find difficult or even impossible.

In summary, our contributions are as follows:
\begin{enumerate}
\item Our intelligent PDE system provides a new way to design
PDEs for computer vision. Based on this framework, we can design particular PDEs for
different vision tasks using different sources of training images~\footnote{Similar idea also appeared in
\cite{Liu:2010:LPDE}. But in that work, the authors only train
special PDEs involving the curvature operator for basic image
restoration tasks. In contrast, our work here proposes a more
unified and elegant framework for more problems in computer
vision.}.
This may be very difficult for traditional PDE design methods.
However, we would like to remind the readers that we have no
intention to beat all the existing approaches for each task, because
these approaches have been carefully and specially tuned for the
task.

\item We propose a general data-based optimal control framework
for training the PDE system. Fed with pairs of input and output
images, the proposed PDE-constrained optimal control training
model can automatically learn the combination coefficients in the
PDE system. Unlike previous design methods, our approach requires
much less human wits and can solve more difficult problems in
computer vision.
\end{enumerate}

The rest of the paper is structured as follows. We first introduce
in Section~\ref{sec:pde} the general intelligent PDE system. In
Section~\ref{sec:opt} we utilize the PDE-constrained optimal control
technique as the training framework for our intelligent PDE system.
Then in Section~\ref{sec:exp} we evaluate our intelligent PDE system
with optimal control training framework by a series of computer
vision and image processing problems. Finally, we give concluding
remarks and a discussion on the future work in
Section~\ref{sec:con}.

%\subsection{Subsection Heading Here}
%Subsection text here.

% needed in second column of first page if using \IEEEpubid
%\IEEEpubidadjcol

\section{General PDE System for Computer Vision}\label{sec:pde}

\subsection{Electromagnetic Field vs. Image Evolution}
Electromagnetism is the force that causes the interaction among
electrically charged particles. The areas in which electromagnetic
interaction happens are called the electromagnetic fields. But in
physics, electrically charged objects were first thought to produce
two types of fields associated with their charge property: An
electric field and a magnetic field. Over time, it was realized that
the electric and magnetic fields are better thought of as two parts
of a greater whole -- the electromagnetic field
\cite{Cheng:1989:EM}. It affects the behavior of charged objects in
the vicinity of the field. The electromagnetic field extends
infinitely throughout space and describes the electromagnetic
interaction. The theoretical implications of electromagnetism also
led to the development of special relativity by Albert Einstein in
1905.

In this paper, inspired by this fundamental force of nature, we
consider the image evolution in a similar way. For a target image
signal $u(t)$, different from most traditional ways, which only
consider the evolution in the image scale space, we define a
companion signal $v(t)$ named the indicator signal. It changes
with time and guides the evolution of $u(t)$ by collecting large
scale information in the image. In this way, these two signals
evolve in two coupled scale spaces.

\subsection{The Intelligent PDE System}
Similar to Maxwell's equations \cite{Cheng:1989:EM}, which are a
set of PDEs describing how the electric and magnetic fields relate
to their sources and how they develop with time, we propose a
general PDE system for the evolution of our coupled signals.

The space of all PDEs is infinitely dimensional. To find the right
form, we start with the properties that our PDE system should have,
in order to narrow down the search space. We notice that
translationally and rotationally invariant properties are very
important for computer vision, i.e., in most vision tasks, when the
input image is translated or rotated, the output image is also
translated or rotated by the same amount. So we require that our PDE
system is translationally and rotationally invariant. According to
the differential invariant theory \cite{Olver:93:applications}, the
form of our PDEs must be functions of the fundamental differential
invariants under the group of translation and rotation. The
fundamental differential invariants are invariant under translation
and rotation and other invariants can be written as their functions.
We list those up to second order in Table~\ref{tab:inv}, where some
notations can be found in Table~\ref{tab:not}. In the sequel, we
shall use $\{\mbox{inv}_j(u,v)\}_{j=0}^{16}$ to refer to them in
order. Note that those invariants are ordered with $u$ going before
$v$. We may reorder them with $v$ going before $u$. In this case,
the $j$-th invariant will be referred to as $\mbox{inv}_j(v,u)$. So
the simplest choice of our general PDE system is the linear
combination of the differential invariants, leading to the following
form:
\begin{equation}
\left\{\begin{array}{lr}
\frac{\partial u}{\partial t} - F(u,v,\{a_j\}_{j=0}^{16}) = 0, & (x,y,t)\in Q,\\
u(x,y,t)=0, & (x,y,t)\in \Gamma,\\
u|_{t=0}=f_u, & (x,y) \in \Omega,\\
\frac{\partial v}{\partial t} - F(v,u,\{b_j\}_{j=0}^{16}) = 0, & (x,y,t)\in Q,\\
v(x,y,t)=0, & (x,y,t)\in \Gamma,\\
v|_{t=0}=f_v, & (x,y) \in \Omega,\\
\end{array} \right.\\\label{eq:lpde}
\end{equation}
where
\begin{equation}
\begin{array}{l}
F(u,v,\{a_j\}_{j=0}^{16})=\sum_{j=0}^{16}a_j(t)\mbox{inv}_j(u, v),\\
F(v,u,\{b_j\}_{j=0}^{16})=\sum_{j=0}^{16}b_j(t)\mbox{inv}_j(v, u),
\end{array}
\end{equation}
$\Omega$ is the rectangular region occupied by the input image $I$,
$T$ is the time that the PDE system finishes the visual information
processing and outputs the results, and $f_u$ and $f_v$ are the
initial functions of $u$ and $v$, respectively. The meaning of other
notations in (\ref{eq:lpde}) can be found in Table~\ref{tab:not}.
For computational issues and the ease of mathematical deduction, $I$
will be padded with zeros of several pixels width around it. As we
can change the unit of time, it is harmless to fix $T = 1$.
$\{a_j(t)\}_{j=0}^{16}$ and $\{b_j(t)\}_{j=0}^{16}$ are sets of
functions defined on $Q$ that are used to control the evolution of
$u$ and $v$, respectively. As $\nabla u$ and $\mathbf{H}_{u}$ change
to $\mathbf{R}\nabla u$ and
$\mathbf{R}\mathbf{H}_{u}\mathbf{R}^{T}$, respectively, when the
image is rotated by a matrix $\mathbf{R}$, it is easy to check the
rotational invariance of those quantities. So the PDE system
(\ref{eq:lpde}) is rotationally invariant. Furthermore, the
following proposition implies that the control functions $a_j(t)$
and $b_j(t)$ can be functions of $t$ only.

\begin{proposition}\label{thm:inv}
Suppose the PDE system (\ref{eq:lpde}) is translationally
invariant, then the control functions $\{a_j\}_{j=0}^{16}$ and
$\{b_j\}_{j=0}^{16}$ must be independent of $(x,y)$.
\end{proposition}
The proof of Proposition~\ref{thm:inv} is presented in \ref{app:invprof}.

%\begin{table*}
%\caption{}\label{tab:inv}
%\end{table*}

\begin{table*}
\begin{center}
\caption{Translationally and rotationally invariant fundamental
differential invariants up to the second order.}\label{tab:inv}
\begin{tabular}{|c|c|l|l|}
\hline

$j$ & $\mbox{inv}_j(u,v)$\\\hline

0,1,2 & $1$, $v$, $u$
\\\hline

3,4 & $\|\nabla v\|^2 = v_x^2 + v_y^2$,  $\|\nabla u\|^2 =
u_x^2+u_y^2$ \\\hline

5 & $(\nabla v)^T\nabla u = v_xu_x + v_yu_y$\\\hline

6,7 & $\mbox{tr}(\mathbf{H}_v) = v_{xx}+v_{yy}$,
$\mbox{tr}(\mathbf{H}_u) = u_{xx}+u_{yy}$
\\\hline

8 & $(\nabla v)^T\mathbf{H}_v\nabla v = v_x^2v_{xx} + 2v_xv_yv_{xy}
+ v_y^2v_{yy}$ \\\hline

9 & $(\nabla v)^T\mathbf{H}_u\nabla v = v_x^2u_{xx} + 2v_xv_yu_{xy}
+ v_y^2u_{yy}$  \\\hline

10 & $(\nabla v)^T\mathbf{H}_v\nabla u = v_xu_xv_{xx} + (v_xu_y +
v_yu_x)v_{xy} + v_yu_yv_{yy}$ \\\hline

11 & $(\nabla v)^T\mathbf{H}_u\nabla u = v_xu_xu_{xx} + (v_xu_y +
v_yu_x)u_{xy} + v_yu_yu_{yy}$ \\\hline

12 & $(\nabla u)^T\mathbf{H}_v\nabla u = u_x^2v_{xx} + 2u_xu_yv_{xy}
+ u_y^2v_{yy}$ \\\hline

13 & $(\nabla u)^T\mathbf{H}_u\nabla u = u_x^2u_{xx} + 2u_xu_yu_{xy}
+ u_y^2u_{yy}$ \\\hline

14 & $\mbox{tr}(\mathbf{H}_v^2) = v_{xx}^2 + 2v_{xy}^2 + v_{yy}^2$
\\\hline

15 & $\mbox{tr}(\mathbf{H}_v\mathbf{H}_u) = v_{xx}u_{xx} +
2v_{xy}u_{xy} + v_{yy}u_{yy}$ \\\hline

16 & $\mbox{tr}(\mathbf{H}_u^2) = u_x^2 + 2u_{xy}^2 + u_y^2$
\\\hline
\end{tabular}
\end{center}
\end{table*}

%\begin{table*}
%\begin{center}
%\caption{}\label{tab:not}
%\end{center}
%\end{table*}

\begin{table*}
\begin{center}
\caption{Notations.\label{tab:not}}
\begin{tabular}{|c|c|c|c|}
\hline $\Omega$ & An open bounded region in $\mathbb{R}^{2}$ &
$\partial\Omega$ & Boundary of $\Omega$\\\hline

$(x,y)$ & $(x,y) \in \Omega$, spatial variable & $t$ & $t \in
(0,T)$, temporal variable\\\hline

$Q$ & $\Omega \times (0,T)$ & $\Gamma$ & $\partial \Omega \times (0,
T)$\\\hline

$|\cdot|$ & The area of a region & $\mathbf{X}^{T}$ & Transpose of
matrix (or vector)
\\\hline

$\|\cdot\|$ & $L^2$ norm & $\mbox{tr}(\cdot)$ & Trace of matrix
\\\hline

%$u(x, y)$, $v(x, y)$ & Image and indicator, respectively &
%$\mathbf{F}^{T}$ & Transposition of $\mathbf{F}$\\\hline

$\nabla u$ & Gradient of $u$ & $\mathbf{H}_u$ & Hessian of $u$
\\\hline

$\wp$ & \multicolumn{3}{c|}{$\wp =
\{(0,0),(0,1),(1,0),(0,2),(1,1),(2,0)\}$, index set for partial
differentiation}\\\hline

%$\kappa (u)$ & \multicolumn{3}{c|}{$\kappa (u)=\mbox{div}
%\left(\frac{\nabla u}{\|\nabla u\|}\right)$, mean curvature of
%$u$}\\\hline

%$\mathbf{inv}(f)$ & \multicolumn{3}{c|}{Functional differential
%invariants of $f$}\\\hline

%$\mbox{sign}(u)$ &
%\multicolumn{3}{c|}{$\mbox{sign}(u)=\frac{u}{|u|}$, signum
%functional of $f$}\\\hline

\end{tabular}
\end{center}
\end{table*}

\section{Training the PDE System via Data-based Optimal Control}\label{sec:opt}
In this section, we propose a data-based optimal control framework
to train the intelligent PDE system for particular vision tasks.

\subsection{The Objective Functional}
Given the forms of PDEs shown in (\ref{eq:lpde}), we have to
determine the coefficient functions $a_j(t)$ and $b_j(t)$. We may
prepare training samples $\{(I_m, O_m)\}_{m=1}^M$, where $I_m$ is
the input image and $O_m$ is the expected output image, and compute
the coefficient functions that minimize the following functional:
\begin{equation}
\begin{array}{l}
J(\{u_m\}_{m=1}^M,\{a_j\}_{j=0}^{16},\{b_j\}_{j=0}^{16})\\
=\frac{1}{2}\sum_{m=1}^M\int_{\Omega}[u_m(x,y,1)-O_m ]^2\d\Omega\\
+ \frac{1}{2}\sum_{j=0}^{16}\lambda_j\int_{0}^{1}a_j^2(t)\d t +
\frac{1}{2}\sum_{j=0}^{16}\mu_j\int_{0}^{1}b_j^2(t)\d t,
\end{array}
\end{equation}
where $u_m(x,y,1)$ is the output image at time $t=1$ computed from
(\ref{eq:lpde}) when the input image is $I_m$, and $\lambda_i$ and
$\mu_i$ are positive weighting parameters. The first term requires
that the final output of our PDE system be close to the ground
truth. The second and the third terms are for regularization so that
the optimal control problem is well posed, as there may be multiple
minimizers for the first term.

\subsection{Solving the Optimal Control Problem}
Then we have the following optimal control problem with PDE
constraints:
\begin{equation}
 \begin{array}{l}
 \mathop{\arg\min}\limits_{\{a_j\}_{j=0}^{16}, \{b_j\}_{j=0}^{16}} J(\{u_m\}_{m=1}^M,\{a_j\}_{j=0}^{16},\{b_j\}_{j=0}^{16})\\
 s.t.
 \left\{\begin{array}{lr}
\frac{\partial u}{\partial t} - F(u,v,\{a_j\}_{j=0}^{16}) = 0, & (x,y,t)\in Q,\\
u(x,y,t)=0, & (x,y,t)\in \Gamma,\\
u|_{t=0}=f_u, & (x,y) \in \Omega,\\
\frac{\partial v}{\partial t} - F(v,u,\{b_j\}_{j=0}^{16}) = 0, & (x,y,t)\in Q,\\
v(x,y,t)=0, & (x,y,t)\in \Gamma,\\
v|_{t=0}=f_v, & (x,y) \in \Omega.\\
\end{array} \right. \\
 \end{array}\label{eq:opc}
\end{equation}
By introducing the adjoint equation of (\ref{eq:opc}), the G\^ateaux
derivative of $J$ can be computed and consequently, the (local)
optimal $\{a_j\}_{j=0}^{16}$ and $\{b_j\}_{j=0}^{16}$ can be
computed via gradient-based algorithms (e.g., conjugate gradient).
Here, we give the adjoint equation and G\^ateaux derivative
directly:

\subsubsection{Adjoint Equation}
\begin{equation}
\left\{\begin{array}{lr}
\frac{\partial \varphi_m}{\partial t} + E(u_m,v_m,\varphi_m,\phi_m) =0, & (x,y,t)\in Q,\\
\varphi_m=0, & (x,y,t) \in \Gamma,\\
\varphi_m|_{t=1} = O_m-u_m(1), & (x,y) \in \Omega,\\
\frac{\partial \phi_m}{\partial t} + E(v_m,u_m,\phi_m,\varphi_m) = 0, & (x,y,t)\in Q,\\
\phi_m=0,  & (x,y,t) \in \Gamma,\\
\phi_m|_{t=1} = 0, & (x,y) \in \Omega,\\
\end{array}\right.\label{eq:ad}
\end{equation}
where
\begin{equation}
\begin{array}{l}
E(u_m,v_m,\varphi_m,\phi_m)\\
=\sum\limits_{(p, q) \in \wp} (-1)^{p+q} \frac{\partial^{p+q}
(\sigma_{pq}(u_m)\varphi_m+\sigma_{pq}(v_m)\phi_m)}{\partial x^p
\partial y^q},\\
E(v_m,u_m,\phi_m,\varphi_m)\\
= \sum\limits_{(p, q) \in \wp} (-1)^{p+q} \frac{\partial^{p+q}
(\sigma_{pq}(u_m)\varphi_m+\sigma_{pq}(v_m)\phi_m)}{\partial x^p
\partial y^q},\\
\sigma_{pq}(u)=\frac{\partial
F(u)}{\partial{u_{pq}}}=\sum_{j=0}^{16}a_j\frac{\partial
\mbox{inv}_j(u, v)}{\partial{u_{pq}}}, \ u_{pq}=\frac{\partial^{p+q}
u} {\partial x^p \partial y^q},\\
\sigma_{pq}(v)=\frac{\partial
F(v)}{\partial{v_{pq}}}=\sum_{j=0}^{16}b_j\frac{\partial
\mbox{inv}_j(v, u)}{\partial{v_{pq}}}, \ v_{pq}=\frac{\partial^{p+q}
v} {\partial x^p \partial y^q}.
\end{array}
\end{equation}
\subsubsection{G\^ateaux Derivative of the Functional}
With the help of the adjoint equation, at each iteration the
derivative of $J$ with respect to $a_j(t)$ and $b_j(t)$ are as
follows:
\begin{equation}
\begin{array}{l}
\frac{\partial J}{\partial
a_j}=\lambda_ja_j-\sum\limits_{m=1}^{M}\int_{\Omega}\varphi_{m}\mbox{inv}_j(u_m,
v_m)\d\Omega,
\quad j=0,...,16,\\
\frac{\partial J}{\partial
b_j}=\mu_jb_j-\sum\limits_{m=1}^{M}\int_{\Omega}\phi_{m}\mbox{inv}_j(v_m,
u_m)\d\Omega, \quad j=0,...,16,\label{eq:gradient}
\end{array}
\end{equation}
where the adjoint functions $\varphi_{m}$ and $\phi_{m}$ are the
solutions to (\ref{eq:ad}).

\subsubsection{Initialization}
Good initialization increases the approximation accuracy of the
learnt PDEs. In our current implementation, we simply set the
initial functions of $u$ and $v$ as the input image:
$$
u_{m}(x,y,0) = v_{m}(x,y,0) = I_m(x,y), \ m = 1,2,...,M.
$$
Then we employ a heuristic method to initialize the control
functions. At each time step, $\frac{\partial u_m}{\partial t}$ is
expected to be $d_m(t) = \frac{O_m - u_m(t)}{1-t}$ so that $u_m(t)$
moves towards the expected output $O_m$ and by the form of
(\ref{eq:lpde}) we may solve $\{a_j(t)\}_{j=0}^{16}$ such that
\begin{equation}
\sum_{m=1}^M\int_{\Omega}[F(u_m, v_m, \{a_j(t)\}_{j=0}^{16}) -
d_m(t)]^2\d\Omega\label{eq:init}
\end{equation}
is minimized\footnote{It is to minimize the difference between the
left and the right hand sides of (\ref{eq:lpde}).}. In this way,
we initialize $a_j(t)$ successively in time while fixing $b_j(t) =
0,j = 0,1,...,16$.

\subsubsection{Finite Difference Method for Numerical Solution}
To solve the intelligent PDE system numerically, we design a
finite difference scheme \cite{Jain:77:partial} for the PDEs. We
discretize the PDEs, i.e. replace the derivatives $\frac{\partial
f}{\partial t}$, $\frac{\partial f}{\partial x}$ and
$\frac{\partial^2 f}{\partial x^2}$ with finite differences as
follows:
\begin{equation}
\left\{\begin{array}{lr}
\frac{\partial f}{\partial t} = \frac{f(t + \Delta t) - f(t)}{\Delta t},\\
\frac{\partial f}{\partial x} = \frac{f(x + 1) - f(x)}{2},\\
\frac{\partial^2 f}{\partial x^2} = f(x - 1) - 2f(x) + f(x + 1).\\
\end{array} \right.\\\label{eq:lpde_fd}
\end{equation}
The discrete forms of $\frac{\partial f}{\partial y}$,
$\frac{\partial^2 f}{\partial y^2}$ and $\frac{\partial^2
f}{\partial x \partial y}$ can be defined similarly. In addition,
we discretize the integrations as
\begin{equation}
\left\{\begin{array}{lr}
\int_{\Omega}f(x,y)\d\Omega = \frac{1}{N}\sum_{\Omega}f(x,y),\\
\int_{0}^{t}f(t)\d t = \Delta t \sum_{i=0}^{T_m} f(i\cdot\Delta t),
\end{array} \right.\\\label{eq:adm_fd}
\end{equation}
where $N$ is the number of pixels in the spatial area, $\Delta t$
is a properly chosen time step size and $T_m=\lfloor\frac{1}{\Delta
t}+0.5\rfloor$ is the index of the expected output time. Then we
use an explicit scheme to compute the numerical solutions.

\subsection{The Optimal-Control-Based Training Framework}
We now summarize in Algorithm~\ref{alg:L-PDE} the data-based optimal
control training framework for the intelligent PDE system. After the
PDE system is learnt, it can be applied to new test images by
solving (\ref{eq:lpde}), whose inputs $f_u$ and $f_v$ are both the
test image and the solution $u(t)|_{t=1}$ is the desired output
image.
\begin{algorithm}[h]
\caption{\bf (Data-based optimal control framework for training the PDE
system)}
\begin{algorithmic}[1]
\REQUIRE Training image pairs $\{(I_m, O_m)\}_{m=1}^{M}$. \STATE
Initialize $a_j(t)$, $t = 0,\Delta t,...,1-\Delta t$, by
minimizing (\ref{eq:init}) and fix $b_j(t)=0$, $j=0,1,...,16$.
\WHILE{not converged} \STATE Compute $\frac{\partial J}{\partial
a_j}$ and $\frac{\partial J}{\partial b_j}$, $j=0,...,16$, using
(\ref{eq:gradient}). \STATE Decide the search direction using the
conjugate gradient method \cite{Stoer:98:numerical}. \STATE
Perform golden search along the search direction and update
$a_j(t)$ and $b_j(t)$, $j=0,...,16$. \ENDWHILE \ENSURE The
coefficient functions $\{a_j(t)\}_{j=0}^{16}$ and
$\{b_j(t)\}_{j=0}^{16}$.
\end{algorithmic}
\label{alg:L-PDE}
\end{algorithm}

\section{Experimental Results}\label{sec:exp}
In this section, we apply our data-based optimal control framework
to learn PDEs for four groups of basic computer vision problems:
Natural image denoising, edge detection, blurring and deburring, and
image segmentation and object detection. As our goal is to show that
the data-based optimal control framework could be a new approach for
designing PDEs and an effective regressor for many computer vision
tasks, \emph{NOT} to propose better algorithms for these tasks, we
are not going to fine tune our PDEs and then compare it with the
state-of-the-art algorithms in every task.

%The curves of the learnt
%coefficients for all experiments are shown in
%Fig.~\ref{fig:denoise_a}, Fig~\ref{fig:edge_a},
%Fig.~\ref{fig:blur_a}, Fig.~\ref{fig:seg_a} and
%Fig.~\ref{fig:det_a}, respectively.

\subsection{Learning from Ground Truth: Natural Image Denoising}
Image denoising is one of the most fundamental low-level vision
problems. For this task, we compare our learnt PDEs with the existing PDE-based
denoising methods, ROF \cite{Rudin:1992:ROF} and TV-$L^1$
\cite{Chan:2005:TVL1}, on images with unknown natural noise. This
task is designed to demonstrate that our method can solve problems
by learning from the ground truth. This is the first advantage of
our data-based optimal control model. We take $240$ images, each
with a size of $150 \times 150$ pixels, of $11$ objects using a
Canon 30D digital camera, setting its ISO to $1600$. For each
object, $30$ images are taken without changing the camera settings
(by fixing the focus, aperture and exposure time) and without
moving the camera position. The average image of them can be
regraded as the noiseless ground truth image. We randomly choose
$8$ objects. For each object we randomly choose $5$ noisy images.
These noisy images and their ground truth images are used to train
the PDE system. Then we compare our learnt PDEs with the
traditional PDEs in \cite{Rudin:1992:ROF} and TV-$L^1$
\cite{Chan:2005:TVL1} on images of the remaining $3$ objects.

\begin{figure*}[htbp]
\centering
\includegraphics[width=0.8\textwidth,
keepaspectratio]{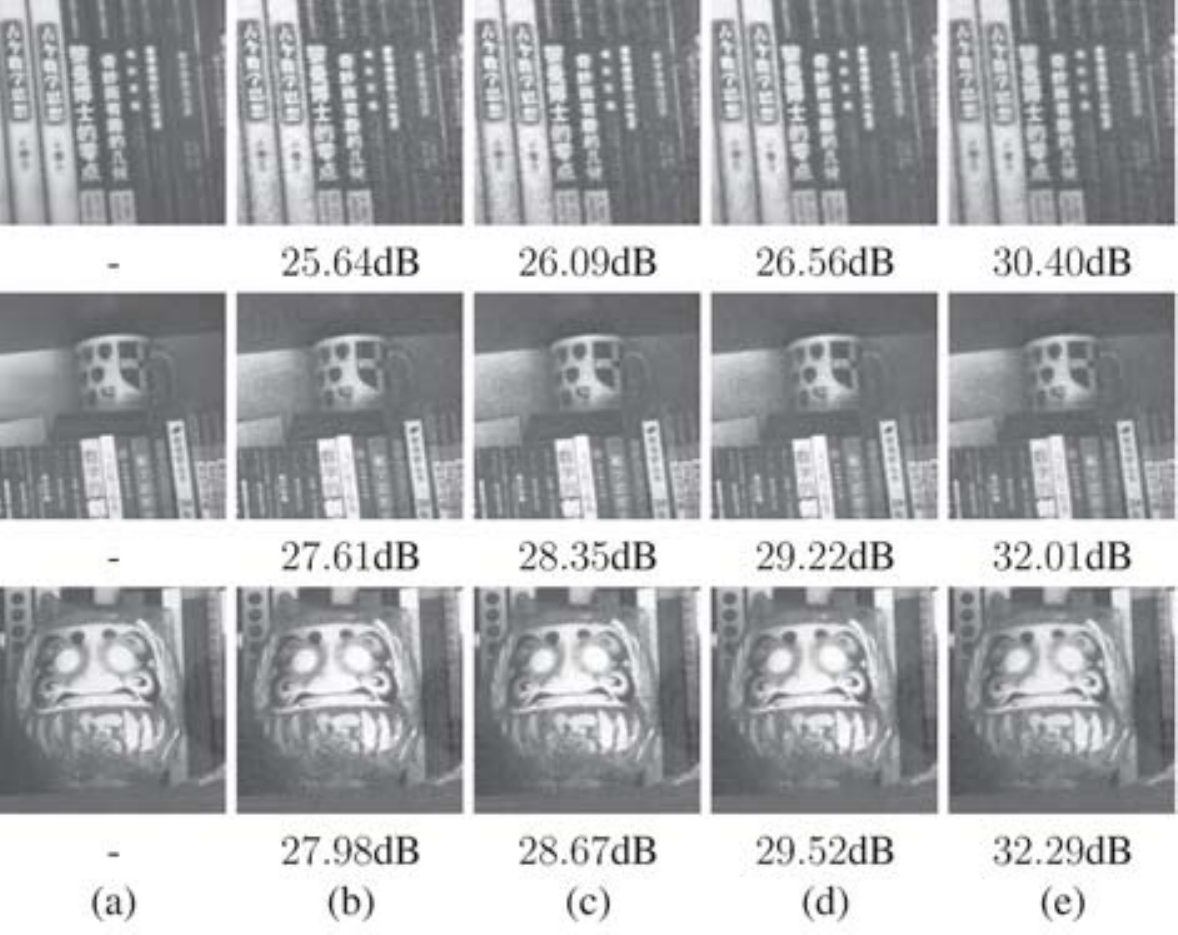}
\caption{The results of denoising images with natural noise. (a)
original noiseless image. (b) noisy image with real noise. (c)-(e)
denoised images using the ROF, TV-$L^1$, and our intelligent PDE
system, respectively. The PSNRs are presented below each
image.}\label{fig:denoise}
\end{figure*}

\begin{figure*}[htbp]
\centering
\includegraphics[width=0.8\textwidth,
keepaspectratio]{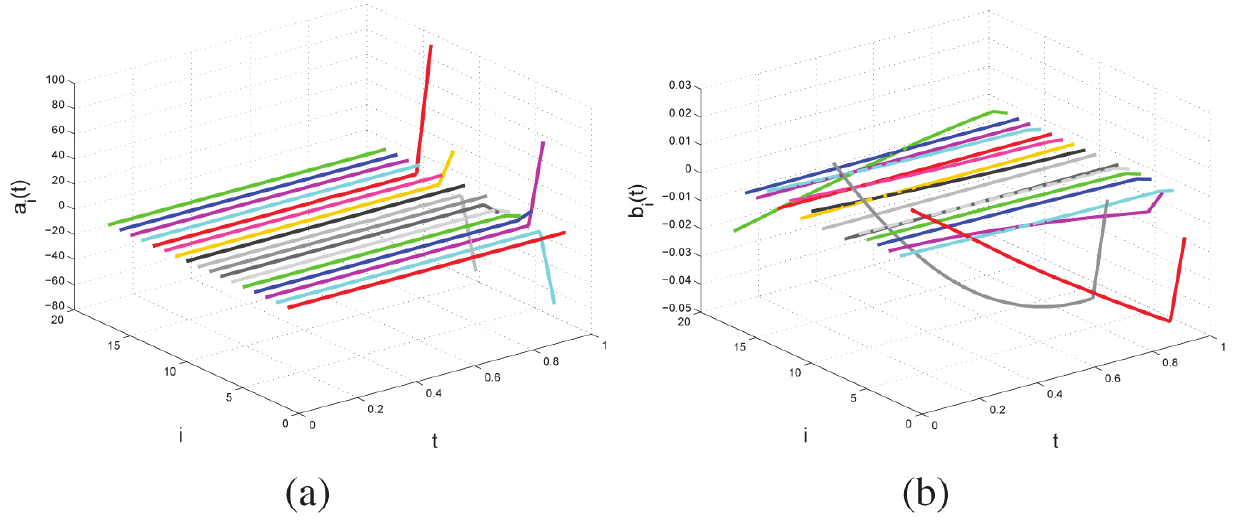}
\caption{Learnt coefficients of the intelligent PDE system for the
natural noise denoising problem. (a) the coefficients
$\{a_j(t)\}_{j=0}^{16}$ for the image evolution. (b) the
coefficients $\{b_j(t)\}_{j=0}^{16}$ for the indicator evolution.}
\label{fig:denoise_a}
\end{figure*}

Fig.~\ref{fig:denoise} shows the comparison results. One can see
that the PSNRs of our intelligent PDEs are dramatically higher
than those of traditional PDEs. This is because our data-based PDE
learning framework can easily adapt to unknown types of noise and
obtain PDE forms to fit for the natural noise well, while most
traditional PDE-based denoising methods were designed under
specific assumptions on the types of noise (e.g., ROF is designed
for Gaussian noise \cite{Rudin:1992:ROF} while TV-$L^1$ is
designed for impulsive noise \cite{Nikolova:2004:TVL1_imno}).
Therefore, they may not fit for unknown types of noise as well as
our intelligent PDEs. The curves of the learnt
coefficients for image denoising are shown in Fig.~\ref{fig:denoise_a}.

%\begin{figure*}[thbp]
%\center
%\includegraphics[width=\columnwidth]{fig2.eps}\\
%\caption{The results of denoising images with natural noise. (a)
%Original noiseless image. (b) Noisy image with real noise. (c)-(e)
%Denoised images using the ROF, TV-$L^1$, and our Intelligent PDE
%models, respectively. The PSNRs are presented below each
%image.}\label{fig:denoise}
%\end{figure*}

%\begin{figure*}[htbp]
%\begin{center}
%\begin{tabular}{c@{\extracolsep{0.2em}}c}
%\includegraphics[width=0.4\textwidth,
%keepaspectratio]{image/eps/a/denoise.eps}
%&\includegraphics[width=0.4\textwidth,
%keepaspectratio]{image/eps/b/denoise.eps}\\
%(a) & (b) \\
%\end{tabular}
%\caption{Learnt coefficients of the intelligent PDE system for the
%natural noise denoising problem. (a) the coefficients
%$\{a_j(t)\}_{j=0}^{16}$ for the image evolution. (b) the
%coefficients $\{b_j(t)\}_{j=0}^{16}$ for the indicator evolution.}
%\label{fig:denoise_a}
%\end{center}
%\end{figure*}

\subsection{Learning from Other Methods: Edge Detection}
The image edge detection task is used to demonstrate that our PDEs
can be learnt from the results of different methods and achieve a
better performance than all of them. This is another advantage of
our data-based optimal control model. For this task, we use three
simple first order edge detectors \cite{Parker:1997:IPCV} (Sobel,
Roberts Cross, and Prewitt) to generate the training data. We
randomly choose $7$ images from the Berkeley image database
\cite{Martin:2001:BSDS} and use the above three detectors to
generate the output images\footnote{This implies that we actually
use a kind of combination of the results from different methods to
train our PDE system.}, together with the input images, to train our
PDE system for edge detection.

\begin{figure*}[htbp]
\centering
\includegraphics[width=0.8\textwidth,
keepaspectratio]{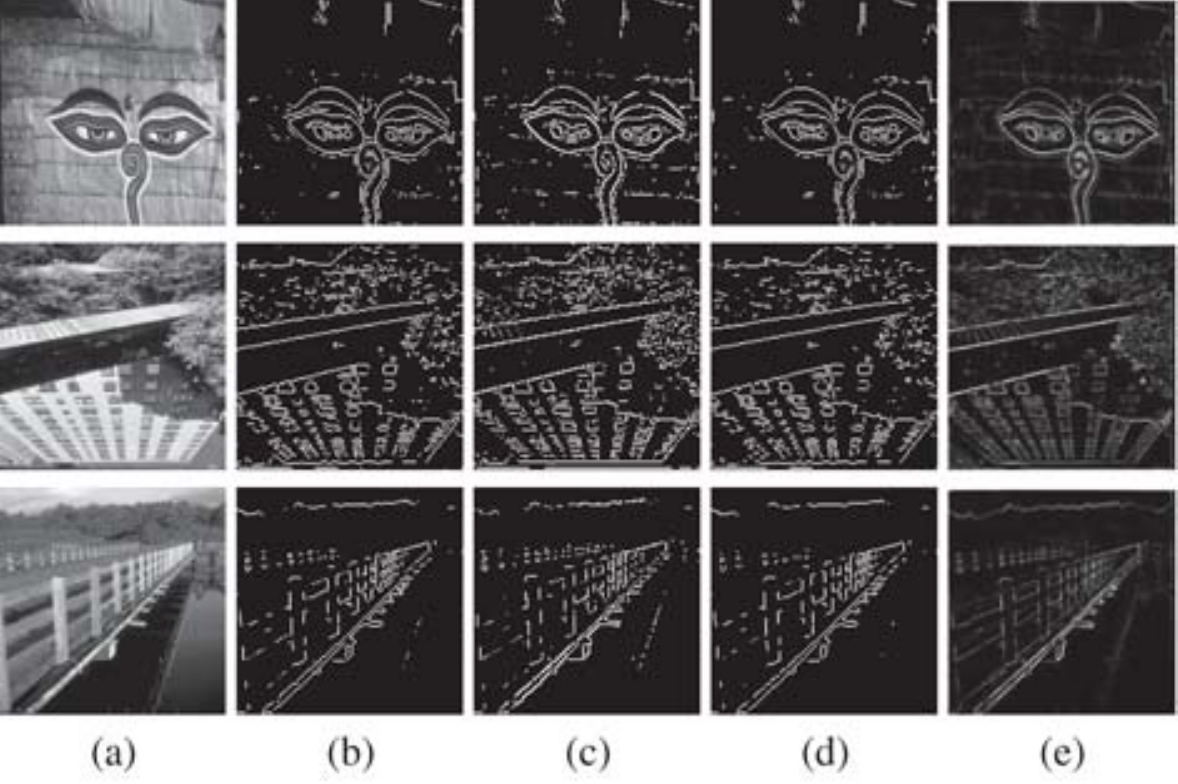}
\caption{The results of edge detection. (a) original image. (b)-(e)
edge detection results using the Sobel, Roberts Cross, Prewitt, and
our intelligent PDE, respectively.}\label{fig:edge}
\end{figure*}

\begin{figure*}[htbp]
\centering
\includegraphics[width=0.8\textwidth,
keepaspectratio]{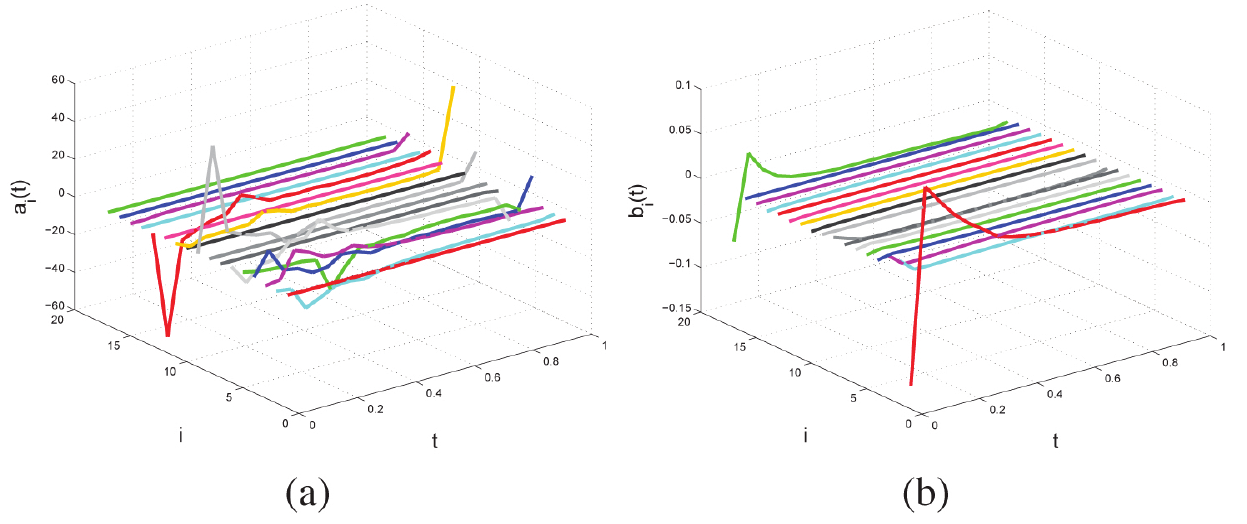}
\caption{Learnt coefficients of the intelligent PDE system for the
edge detection problem. (a) the coefficients
$\{a_j(t)\}_{j=0}^{16}$ for the image evolution. (b) the
coefficients $\{b_j(t)\}_{j=0}^{16}$ for the indicator
evolution.}\label{fig:edge_a}
\end{figure*}

Fig.~\ref{fig:edge} shows part of the edge detection results on
other images in the Berkeley image database. One can see that our
PDEs respond selectively to edges and basically produce visually
significant edges, while the edge maps of other three detectors
are more chaotic. Note that the solution to our PDEs is supposed to
be a more or less smooth function. So one cannot expect that our
PDEs produce an exactly binary edge map. Instead, an approximation
of a binary edge map is produced. The curves of the learnt
coefficients for edge detection are shown in Fig.~\ref{fig:edge_a}.

%\begin{figure*}[htbp]
%\begin{center}
%\begin{tabular}{c@{\extracolsep{0.2em}}c}
%\includegraphics[width=0.4\textwidth,
%keepaspectratio]{image/eps/a/edge.eps}
%&\includegraphics[width=0.4\textwidth,
%keepaspectratio]{image/eps/b/edge.eps}\\
%(a) & (b)\\
%\end{tabular}
%\end{center}
%\caption{Learnt coefficients of the intelligent PDE system for the
%edge detection problem. (a) the coefficients
%$\{a_j(t)\}_{j=0}^{16}$ for the image evolution. (b) the
%coefficients $\{b_j(t)\}_{j=0}^{16}$ for the indicator
%evolution.}\label{fig:edge_a}
%\end{figure*}

\subsection{Learning to Solve Both Primal and Inverse Problems: Blurring and Deblurring}
The traditional PDEs for solving different problems are usually of
very different appearance. The task of solving both blurring
and deblurring is designed to show that the same form of PDEs can be learnt
to solve both the primal and inverse problems. This is the third advantage of our data-based
optimal control model.

\begin{figure*}[htbp]
\centering
\includegraphics[width=0.8\textwidth,
keepaspectratio]{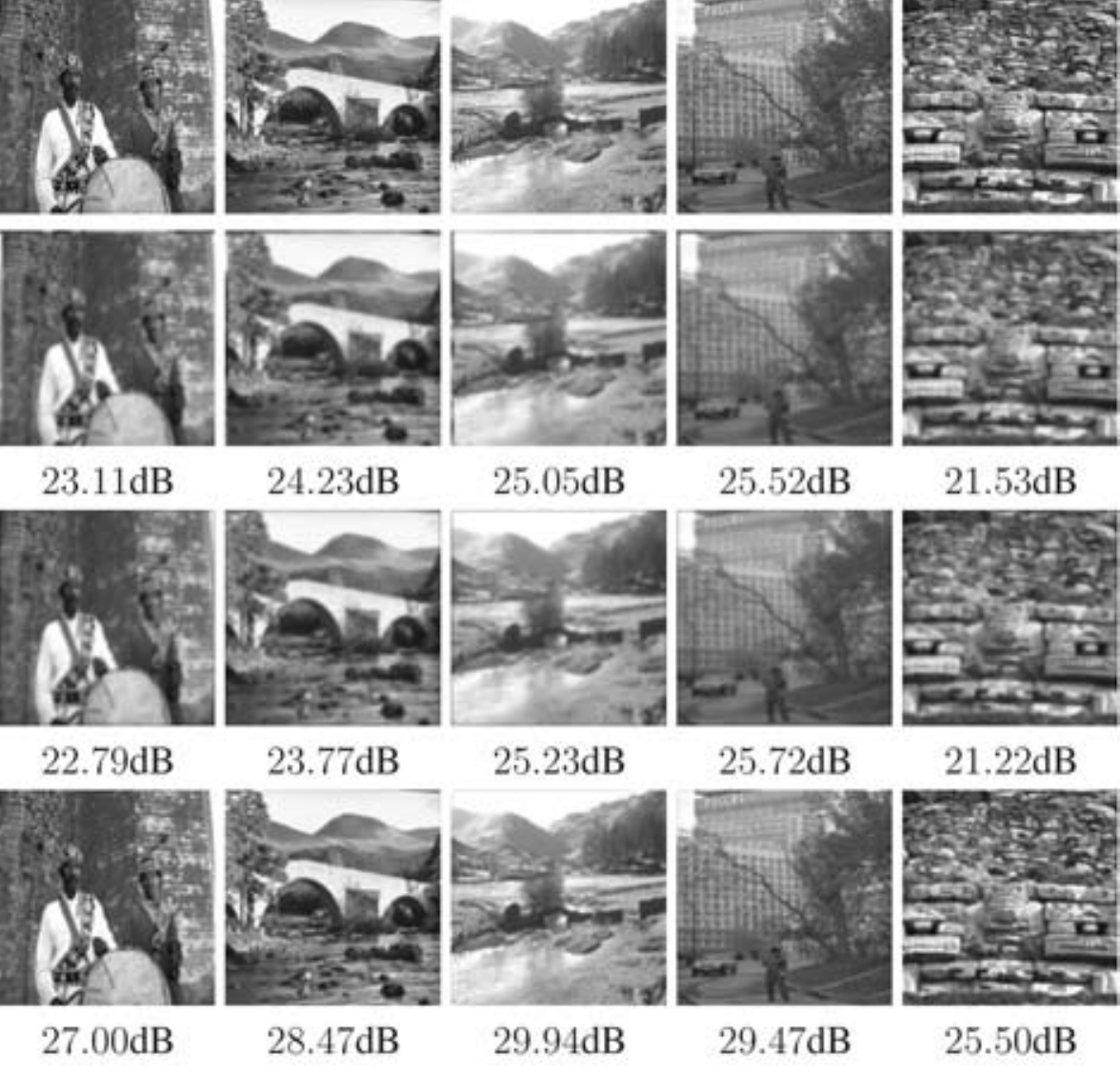}
\caption{The results of image blurring and deblurring. The top row
are the original images. The second row are the blurring results
of a Gaussian kernel. The third row are the blurring results of
our Intelligent PDE. The bottom row are the deblurring results of
our Intelligent PDE. The PSNRs are presented below each
image.}\label{fig:blur}
\end{figure*}

\begin{figure*}[htbp]
\centering
\includegraphics[width=0.8\textwidth,
keepaspectratio]{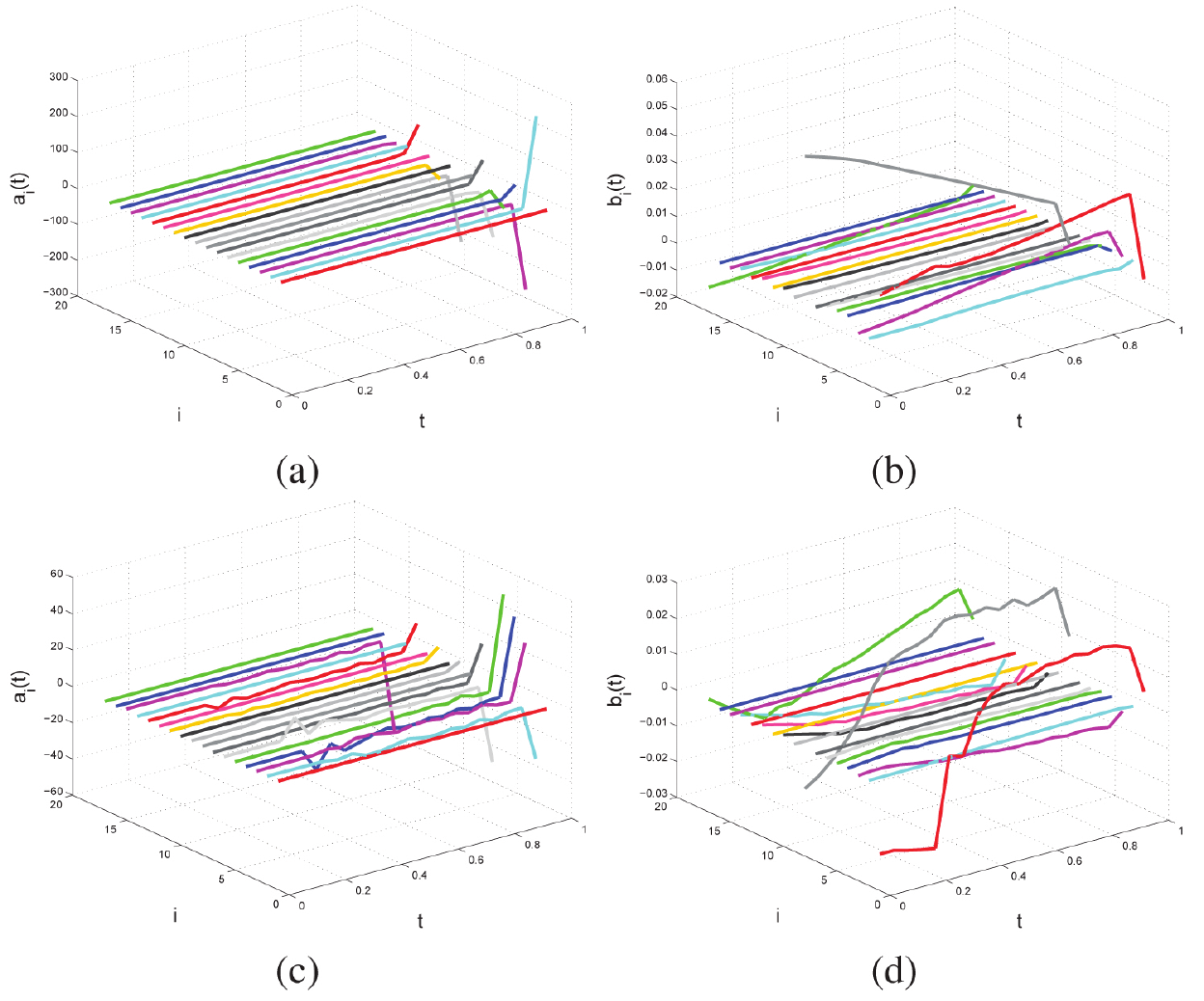}
\caption{Learnt coefficients of the intelligent PDE system for
image blurring (top row) and deblurring (bottom row) problems. (a)
and (c) are the coefficients $\{a_j(t)\}_{j=0}^{16}$ for the image
evolutions. (b) and (d) are the coefficients
$\{b_j(t)\}_{j=0}^{16}$ for the indicator
evolutions.}\label{fig:blur_a}
\end{figure*}

For the image blurring task (the primal problem), the output image
is the convolution of the input image with a Gaussian kernel. So
we generate the output images by blurring high resolution images
using a Gaussian kernel with $\sigma = 1$. The original images are
used as the input. As shown in the third row of
Fig.~\ref{fig:blur}, the output is nearly identical to the ground
truth (the second row of Fig.~\ref{fig:blur}). For the image
deblurring task (the inverse problem), we just exchange the input
and output images for training. One can see in the bottom row of
Fig.~\ref{fig:blur} that the output is very close to the original
image (first row of Fig.~\ref{fig:blur}). The curves of the learnt
coefficients for image deblurring are shown in Fig.~\ref{fig:blur_a}.

%\begin{figure*}[htbp]
%\begin{center}
%\begin{tabular}{c@{\extracolsep{0.2em}}c}
%\includegraphics[width=0.4\textwidth,
%keepaspectratio]{image/eps/a/blur.eps}
%&\includegraphics[width=0.4\textwidth,
%keepaspectratio]{image/eps/b/blur.eps}\\
%(a) & (b)\\
%\includegraphics[width=0.4\textwidth,
%keepaspectratio]{image/eps/a/deblur.eps}
%&\includegraphics[width=0.4\textwidth,
%keepaspectratio]{image/eps/b/deblur.eps}\\
%(c) & (d)\\
%\end{tabular}
%\end{center}
%\caption{Learnt coefficients of the intelligent PDE system for
%image blurring (top row) and deblurring (bottom row) problems. (a)
%and (c) are the coefficients $\{a_j(t)\}_{j=0}^{16}$ for the image
%evolutions. (b) and (d) are the coefficients
%$\{b_j(t)\}_{j=0}^{16}$ for the indicator
%evolutions.}\label{fig:blur_a}
%\end{figure*}

\subsection{Learning from Humans: Image Segmentation and Object Detection}
Image segmentation and object detection are designed to demonstrate
that our PDE system can learn from the human behavior directly
(learn the segmentation and detection results provided by humans,
e.g., manually segmented masks).

\begin{figure*}[htbp]
\centering
\includegraphics[width=0.8\textwidth,
keepaspectratio]{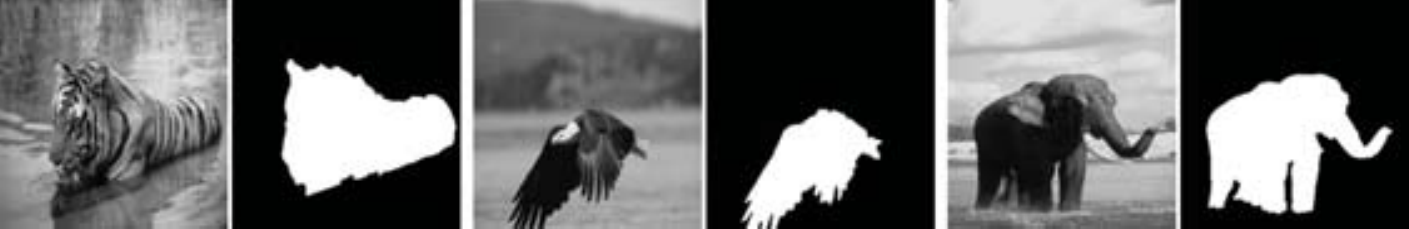}
\caption{Examples of the training images for image segmentation. In
each group of images, on the left is the input image and on the
right is the ground truth output mask.}\label{fig:seg_train}
\end{figure*}

For image segmentation, it is a highly ill-posed problem and there
are many criteria that define the goal of segmentation, e.g.,
breaking an image into regions with similar intensity, color,
texture, or expected shape. As none of the current image
segmentation algorithms can perform object level segmentation well
out of complex backgrounds, we choose to require our PDEs to achieve
a reasonable goal, namely segmenting relatively darker objects
against relatively simple backgrounds, where both the foreground and
the background can be highly textured and simple thresholding cannot
separate them. So we select 60 images from the Corel image database
\cite{Corel} that have relatively darker foregrounds and relatively
simple backgrounds, but the foreground is not of uniformly lower
graylevels than the background, and also prepare the manually
segmented binary masks as the outputs of the training images, where
the black regions are the backgrounds (Fig.~\ref{fig:seg_train}).

%\begin{figure*}[htbp]
%\begin{center}
%\begin{tabular}{c@{\extracolsep{0.1em}}c@{\extracolsep{0.3em}}c@{\extracolsep{0.1em}}c@{\extracolsep{0.3em}}c@{\extracolsep{0.1em}}c}
%\includegraphics[width=0.15\textwidth,
%keepaspectratio]{image/eps/seg/result/seg_8847_train.eps}
%&\includegraphics[width=0.15\textwidth,
%keepaspectratio]{image/eps/seg/result/seg_8847_mask.eps}
%&\includegraphics[width=0.15\textwidth,
%keepaspectratio]{image/eps/seg/result/seg_135099_train.eps}
%&\includegraphics[width=0.15\textwidth,
%keepaspectratio]{image/eps/seg/result/seg_135099_mask.eps}
%&\includegraphics[width=0.15\textwidth,
%keepaspectratio]{image/eps/seg/result/seg_3286_train.eps}
%&\includegraphics[width=0.15\textwidth,
%keepaspectratio]{image/eps/seg/result/seg_3286_mask.eps}
%\end{tabular}
%\end{center}
%\caption{Examples of the training images for image segmentation. In
%each group of images, on the left is the input image and on the
%right is the ground truth output mask.}\label{fig:seg_train}
%\end{figure*}

\begin{figure*}[htbp]
\centering
\includegraphics[width=0.8\textwidth,
keepaspectratio]{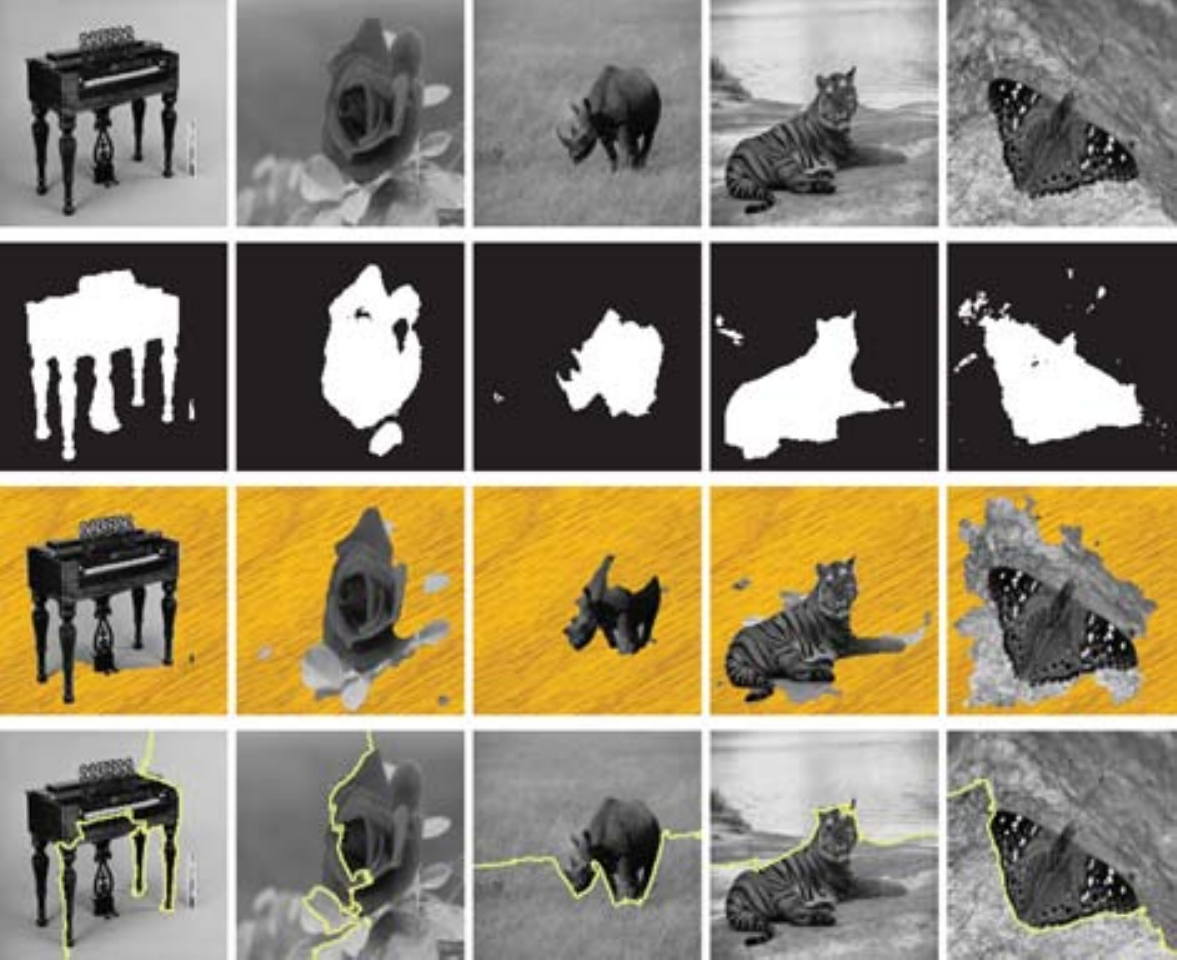}
\caption{The results of image segmentation. The top row are the
segmentation results of active contour \cite{Li:2005:LevelSet} on
the original images. The bottom row are the results obtained by
our Intelligent PDE.}\label{fig:seg}
\end{figure*}

\begin{figure*}[htbp]
\centering
\includegraphics[width=0.8\textwidth,
keepaspectratio]{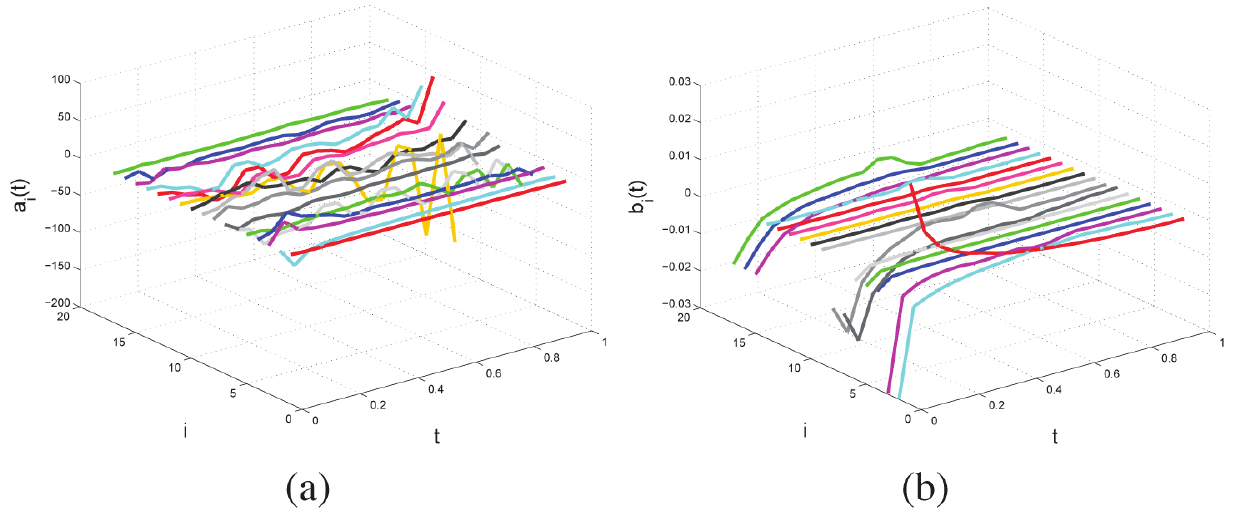}
\caption{Learnt coefficients of the intelligent PDE system for
image segmentation problems. (a) the coefficients
$\{a_j(t)\}_{j=0}^{16}$ for the image evolution. (b) the
coefficients $\{b_j(t)\}_{j=0}^{16}$ for the indicator
evolution.}\label{fig:seg_a}
\end{figure*}

Part of the segmentation results are shown in Fig.~\ref{fig:seg},
where we have set a threshold for the output mask maps of our learnt
PDEs with a constant 0.5. We see that our learnt PDEs produce fairly
good object masks. We also test the active contour method by Li {et
al.} \cite{Li:2005:LevelSet}\footnote{Code available at
http://www.engr.uconn.edu/$\sim$cmli/} and the normalized cut method
\cite{Shi:2000:Ncut}\footnote{Code available at
http://www.cis.upenn.edu/$\sim$jshi/software/}. One can see from
Fig.~\ref{fig:seg} that the active contour method cannot segment
object details due to the smoothness constraint on the object shape
and the normalized cut method cannot produce a closed foreground
region. To provide a quantitative evaluation, we use the
$F$-measures that merge the precision and recall of segmentation:
$$F_\alpha = \frac{(1+\alpha)\cdot recall \cdot precision}{\alpha\cdot precision + recall},
\mbox{ where }$$ $$recall = \frac{|A\cap B|}{|A|},\quad
precision=\frac{|A\cap B|}{|B|},
$$
in which $A$ is the ground truth mask and $B$ is the computed
mask. The most common choice of $\alpha$ is 2. On our test images,
the $F_2$ measures of our PDEs, \cite{Li:2005:LevelSet} and
\cite{Shi:2000:Ncut} are $0.90\pm 0.05$, $0.83\pm 0.16$ and
$0.61\pm 0.20$, respectively. One can see that the performance of
our PDEs is better than theirs, in both visual quality and
quantitative measure. The curves of the learnt
coefficients for image segmentation are shown in Fig.~\ref{fig:seg_a}.

%\begin{figure*}[thbp]
%\center
%\includegraphics[width=\columnwidth]{fig8.eps}\\
%\caption{Examples of the training images for image segmentation. In
%each group of images, on the left is the input image and on the
%right is the ground truth output mask.}\label{fig:seg_train}
%\end{figure*}

%\begin{figure*}[htbp]
%\begin{center}
%\begin{tabular}{c@{\extracolsep{0.2em}}c}
%\includegraphics[width=0.4\textwidth,
%keepaspectratio]{image/eps/a/seg.eps}
%&\includegraphics[width=0.4\textwidth,
%keepaspectratio]{image/eps/b/seg.eps}\\
%(a) & (b)\\
%\end{tabular}
%\end{center}
%\caption{Learnt coefficients of the intelligent PDE system for
%image segmentation problems. (a) the coefficients
%$\{a_j(t)\}_{j=0}^{16}$ for the image evolution. (b) the
%coefficients $\{b_j(t)\}_{j=0}^{16}$ for the indicator
%evolution.}\label{fig:seg_a}
%\end{figure*}

\begin{figure*}[htbp]
\centering
\includegraphics[width=0.8\textwidth,
keepaspectratio]{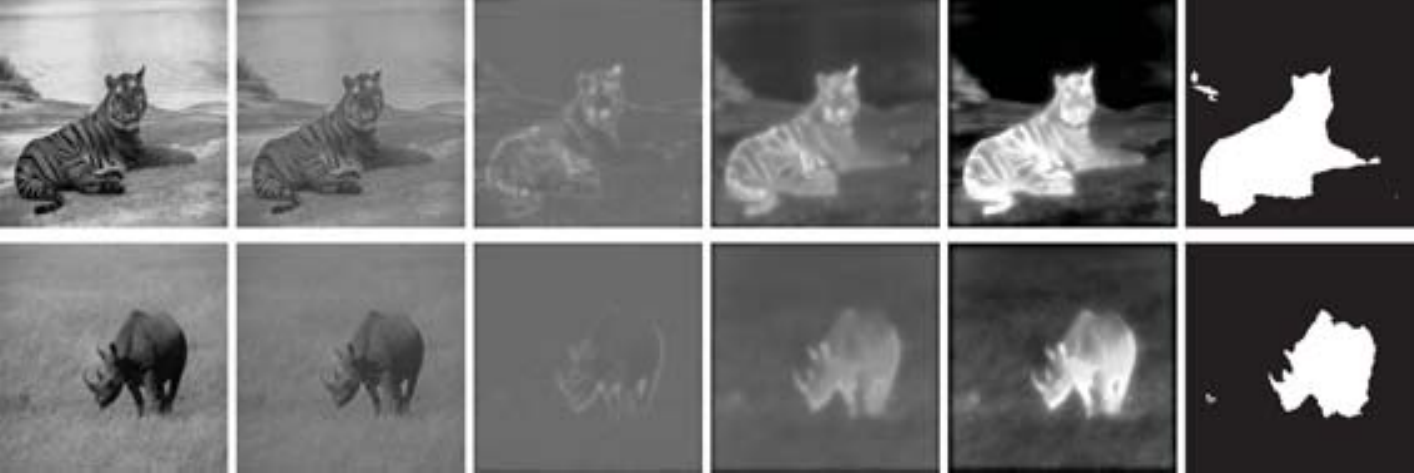}
\caption{The evolution of the mask maps. For each row, the first
image is the input image, the second to the fifth are the mask maps
at time $t=0.25,0.50,0.75,1.0$, respectively, and the last image is
the final mask map with a threshold of 0.5. }\label{fig:evolution}
\end{figure*}

We also present the evolution process of the mask maps across time
(Fig.~\ref{fig:evolution}). One can see that although the foreground
is relatively darker than the background, the PDEs correctly detect
the most salient points/edges and then propagate the information
across the foreground region, resulting in a \emph{brighter} output
region for the foreground.

%\begin{figure*}[thbp]
%\center
%\includegraphics[width=\columnwidth]{fig_evo.eps}\\
%\caption{The evolution of the mask maps. For each row, the first
%image is the input image, the second to the fifth are the mask maps
%at time $t=0.25,0.50,0.75,1.0$, respectively, and the last image is
%the final mask map thresholded at 0.5. }\label{fig:evolution}
%\end{figure*}

Now we apply our intelligent PDEs for a more complex task: Object
detection. Namely, the PDEs should respond strongly to the object of
interest while not responding (or responding much more weakly) if
the object is absent in the image. It should be challenging enough
for one to manually design PDEs to perform such a problem. As a
result, we are unaware of any PDE-based method that can accomplish
this task. The existing PDE-based segmentation algorithms always
output an ``object region'' even if the image does not contain the
object of interest. In contrast, we will show that as desired our
PDEs are able to respond selectively. We choose the ``plane'' data
set in Corel \cite{Corel}. We select $30$ images from this data set
as positive samples and also prepare $30$ images without the object
of interest as negative samples. We also provide their ground truth
object masks\footnote{For the positive samples, we manually segment
binary masks as the output images. For the negative samples, the
ground truth output masks are all-zero images.} in order to complete
the training data.

\begin{figure*}[htbp]
\centering
\includegraphics[width=0.8\textwidth,
keepaspectratio]{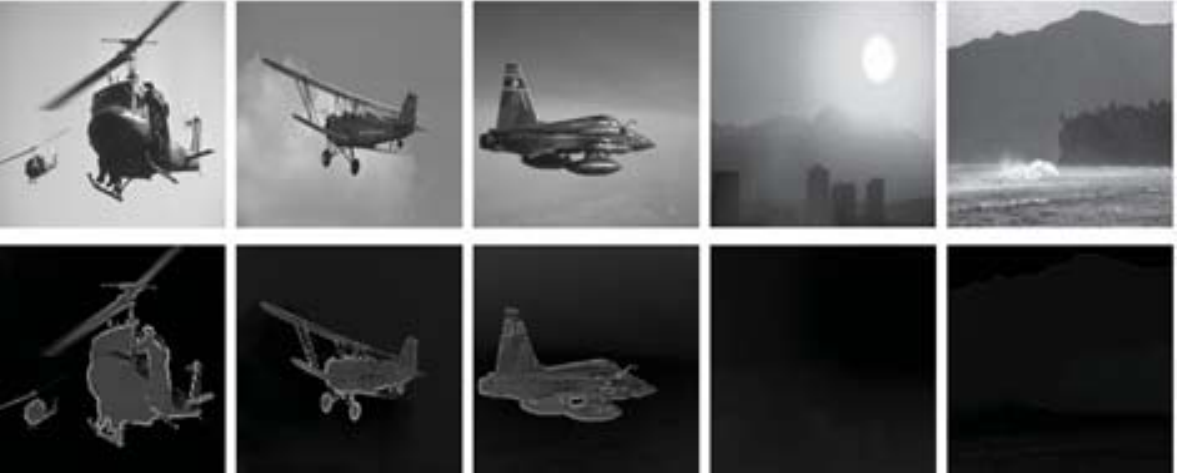}
\caption{The results of ``plane'' detection. The top row are the
original images, the first three are images containing planes and
the last two are without planes. The bottom row are the detection
results of our intelligent PDE for images with and without planes,
respectively.}\label{fig:detection}
\end{figure*}

\begin{figure*}[htbp]
\centering
\includegraphics[width=0.8\textwidth,
keepaspectratio]{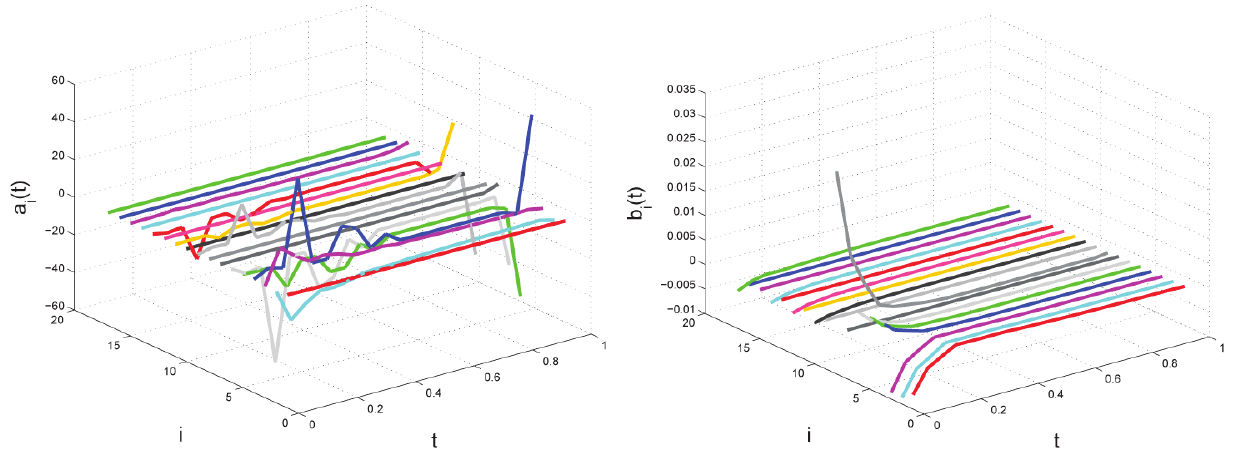}
\caption{Learnt coefficients of the intelligent PDE system for the
plane detection problem. (a) the coefficients
$\{a_j(t)\}_{j=0}^{16}$ for the image evolution. (b) the
coefficients $\{b_j(t)\}_{j=0}^{16}$ for the indicator
evolution.}\label{fig:det_a}
\end{figure*}

In Fig.~\ref{fig:detection}, one can see that our learnt PDEs
respond well to the objects of interest (first three images),
while the response to images without the objects of interest is
relatively low across the whole images (last two images). It seems
that our PDEs automatically identify that the concurrent
high-contrast edges/junctions/corners are the key features of
planes. The above examples show that our learnt PDEs are able to
differentiate the object/non-object regions, without requiring the
user to teach them what features are and what factors to consider. The curves of the learnt
coefficients for object detection are shown in Fig.~\ref{fig:det_a}.

%\begin{figure*}[htbp]
%\begin{center}
%\begin{tabular}{c@{\extracolsep{0.2em}}c@{\extracolsep{0.2em}}c@{\extracolsep{0.2em}}c@{\extracolsep{0.2em}}c}
%\includegraphics[width=0.15\textwidth,
%keepaspectratio]{image/eps/detection/positive/ori/179043.eps}
%&\includegraphics[width=0.15\textwidth,
%keepaspectratio]{image/eps/detection/positive/ori/360098.eps}
%&\includegraphics[width=0.15\textwidth,
%keepaspectratio]{image/eps/detection/positive/ori/37085.eps}
%&\includegraphics[width=0.15\textwidth,
%keepaspectratio]{image/eps/detection/negative/ori/1000.eps}
%&\includegraphics[width=0.15\textwidth,
%keepaspectratio]{image/eps/detection/negative/ori/83003.eps}\\
%\includegraphics[width=0.15\textwidth,
%keepaspectratio]{image/eps/detection/positive/result/179043.eps}
%&\includegraphics[width=0.15\textwidth,
%keepaspectratio]{image/eps/detection/positive/result/360098.eps}
%&\includegraphics[width=0.15\textwidth,
%keepaspectratio]{image/eps/detection/positive/result/37085.eps}
%&\includegraphics[width=0.15\textwidth,
%keepaspectratio]{image/eps/detection/negative/result/1000.eps}
%&\includegraphics[width=0.15\textwidth,
%keepaspectratio]{image/eps/detection/negative/result/83003.eps}\\
%\end{tabular}
%\end{center}
%\caption{The results of ``plane'' detection. The top row are the
%original images, the first three are images containing planes and
%the last two are without planes. The bottom row are the detection
%results of our intelligent PDE for images with and without planes,
%respectively.}\label{fig:detection}
%\end{figure*}

%\begin{figure*}[htbp]
%\begin{center}
%\begin{tabular}{c@{\extracolsep{0.2em}}c}
%\includegraphics[width=0.4\textwidth,
%keepaspectratio]{image/eps/a/detection.eps}
%&\includegraphics[width=0.4\textwidth,
%keepaspectratio]{image/eps/b/detection.eps}\\
%\end{tabular}
%\end{center}
%\caption{Learnt coefficients of the intelligent PDE system for the
%plane detection problem. (a) the coefficients
%$\{a_j(t)\}_{j=0}^{16}$ for the image evolution. (b) the
%coefficients $\{b_j(t)\}_{j=0}^{16}$ for the indicator
%evolution.}\label{fig:det_a}
%\end{figure*}

\section{Conclusion}\label{sec:con}
In this paper, we have presented a framework for using data-based
optimal control to learn PDEs as a general regressor to approximate
the nonlinear mappings of different visual processing tasks. The
experimental results on some computer vision and image processing
problems show that our framework is promising. However, the current
work is still preliminary, so we plan to improve and enrich our work
in the following aspects. First, more theoretical issues should be
addressed for this PDE system. For example, we will try to apply the
Adomian decomposition method \cite{Wazwaz:2009:ADM} to express the
exact analytical solution to (\ref{eq:lpde}) and then analyze its
physical properties. Second, we would like to develop more
computationally efficient numerical algorithms to solve our
PDE-constrained optimal control problem (\ref{eq:opc}). Third, we
will apply our framework to more vision tasks to find out to what
extent it works.

\appendix

\section{Proof of Property~\ref{thm:inv}}\label{app:invprof}
We prove that the coefficients $\{a_j\}_{j=0}^{16}$ and
$\{b_j\}_{j=0}^{16}$ must be independent of $(x,y)$.
\begin{proof}
We prove for $F(u,v,\{a_j\}_{j=0}^{16})$ in (\ref{eq:lpde}) only.
We may rewrite
$$F(u,v,\{a_j\}_{j=0}^{16})=\tilde{F}(u,v,x,y,t).$$
Then it suffices to prove that $\tilde{F}$ is independent of $(x,y)$.

By the definition of translational invariance, when $I(x,y)$ changes
to $I(x-x_0,y-y_0)$ by shifting with a displacement $(x_0,y_0)$,
$u(x,y)$ and $v(x,y)$ will change to $u(x-x_0,y-y_0)$ and
$v(x-x_0,y-y_0)$, respectively. So the pair $u(x-x_0,y-y_0)$ and
$v(x-x_0,y-y_0)$ fulfils (\ref{eq:lpde}), i.e.,
\begin{eqnarray*}
\begin{array}{l}
\displaystyle \frac{\displaystyle \partial u(x-x_0,y-y_0)}{\displaystyle \partial t} \\
\quad - \tilde{F}(u(x-x_0,y-y_0),v(x-x_0,y-y_0),x,y,t) = 0.
\end{array}
\end{eqnarray*}
Next, we replace $(x-x_0,y-y_0)$ in the above equation with $(x,y)$
and have:
$$
\frac{\partial u(x,y)}{\partial t} - \tilde{F}(u(x,y),v(x,y),x+x_0,y+y_0,t) = 0.
$$
On the other hand, the pair $(u(x,y),v(x,y))$ also fulfils
(\ref{eq:lpde}), i.e.,
$$
\frac{\partial u(x,y)}{\partial t} - \tilde{F}(u(x,y),v(x,y),x,y,t) = 0.
$$
Therefore, $\tilde{F}(u,v,x+x_0,y+y_0,t)=\tilde{F}(u,v,x,y,t)$, $\forall (x_0,y_0)$ that
confines the input image inside $\Omega$. So $\tilde{F}$ is
independent of $(x,y)$.
\end{proof}

\section*{Acknowledgment}
This work was partially supported by the grants of the National
Science Foundation of China, No. U0935004 and 60873181.

\end{document}